%% file: main.tex
\documentclass[letterpaper, 10 pt, conference]{ieeeconf}  

\IEEEoverridecommandlockouts                              

\newcommand{\nbname}{\textsc{DexCast}}
\usepackage{wrapfig}
\usepackage{pifont}
\usepackage{array}
\usepackage{hyperref}
\usepackage{url}

\usepackage{booktabs,makecell}
\usepackage{wrapfig}

\usepackage{paracol}
\usepackage{amsmath,amssymb,algorithm,algorithmic}

\usepackage{overpic}
\usepackage{adjustbox}
\usepackage{colortbl}
\usepackage{multirow}
\usepackage{booktabs}
\usepackage{tabularx} 
\usepackage{booktabs,multirow,makecell,array,graphicx}
\definecolor{eccvblue}{rgb}{0.21,0.49,0.74}
\usepackage{threeparttable}
\usepackage{siunitx}
\usepackage{wrapfig}
\usepackage{array}
\usepackage{longtable}
\usepackage{calc}      
\usepackage{caption}   
\usepackage{tikz}
\usetikzlibrary{positioning,arrows.meta}
\usepackage{caption}
\usepackage{subcaption}      
\usepackage{amsmath}
\usepackage{graphicx}
\usepackage{booktabs}
\usepackage[table]{xcolor}   
\definecolor{cellgray}{RGB}{200,200,200}
\usepackage{array}
\usepackage[table]{xcolor}
\usepackage{booktabs}
\usepackage{multirow}
\usepackage{makecell}

\newcolumntype{L}[1]{>{\raggedright\arraybackslash}p{#1}}
\usepackage{cuted}
\usepackage{capt-of}
\usepackage[accsupp]{axessibility}  

\usepackage{orcidlink}
\usepackage{hyperref}
\usepackage{wrapfig}
\usepackage{svg}
\usepackage[nameinlink,capitalize]{cleveref}
\usepackage{xspace}
\title{\textbf{\textit{DexCast}}: Learning Real-World Dexterous Control from Human Videos
}

\author{
\begin{tabular}{@{}c@{}}
{\normalfont\mdseries
Zihan Gao$^{1,*}$ \quad
Yuhong Zhang$^{1,2,*}$ \quad
Mengzhe Wang$^{1,*}$ \quad
Kaisheng Liu$^{3}$ \quad
Shengpeng Li$^{4}$}\\
{\normalfont\mdseries
Siyao Xiao$^{1}$ \quad
Runqing Cheng$^{5}$ \quad
Xiao Lin$^{5}$ \quad
Zhifang Liu$^{1}$ \quad
Haoqian Wang$^{1,\dagger}$}\\[0.3em]
{\normalfont\bfseries
$^{1}$Tsinghua University \quad
$^{2}$TsingTec \quad
$^{3}$Shenzhen University}\\
{\normalfont\bfseries
$^{4}$Nanyang Technological University \quad
$^{5}$Georgia Institute of Technology}\\[0.3em]
{\normalfont\mdseries
\{gaozihanthu, dsyuhong, wangmengzhe076\}@gmail.com}\\
{\normalfont\mdseries\small
$^{*}$Equal contribution. \quad $^{\dagger}$Corresponding author.}
\end{tabular}
}

\begin{document}

\maketitle
\thispagestyle{empty}
\pagestyle{empty}

\crefname{figure}{Figure}{Figures}
\Crefname{figure}{Figure}{Figures}
\maketitle

\begin{strip}
\vspace{-4.2em}
    \centering
    \includegraphics[width=\textwidth]{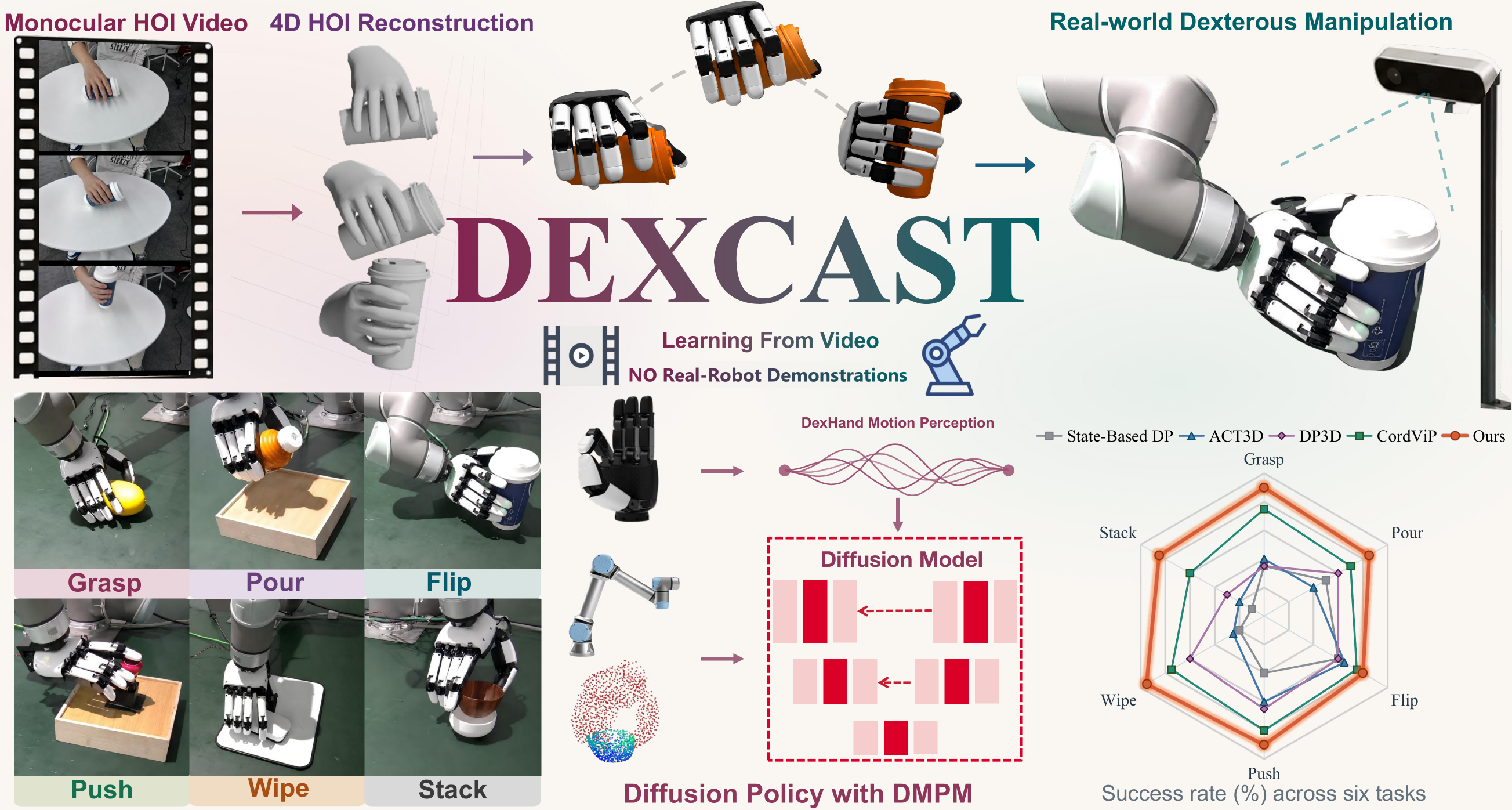}
    \captionof{figure}{\nbname\, establishes a new pathway for learning dexterous skills directly from human video demonstrations. \nbname\, leverages 4D hand–object interaction reconstruction to eliminate the need for labor-intensive demonstration collection and introduces a DexHand Motion Perception Module to enhance dexterous manipulation.}
    \label{fig:teaser}
\end{strip}


\input{sec/0_abstract}    
\input{sec/1_intro}

\input{sec/2_related_work}

\input{sec/3_method}
\input{sec/4_exper}

\input{sec/5_conclusion}


\bibliographystyle{IEEEtran}
\bibliography{main}

\end{document}

%% file: sec/0_abstract.tex
\begin{abstract}
    Recent advances in 3D hand–object interaction (HOI) perception have opened new opportunities for learning dexterous robotic skills from human videos. However, existing dexterous imitation learning methods often rely on costly and manually teleoperated robot demonstrations, requiring substantial human effort. Therefore, we build \nbname, a framework that learns dexterous robotic hand control directly from monocular RGB-D videos by leveraging joint HOI reconstructions, without requiring any teleoperation. Furthermore, how to model high-dimensional robotic hand motion and extract temporal context to inform future action generation remains a key challenge. To this end, we propose a DexHand Motion Perception Module (DMPM) to capture inter-joint coordination and temporal dependencies in compact latent representations that condition our diffusion policy. Extensive real-world experiments demonstrate that DEXCAST achieves exceptional performance across diverse dexterous manipulation tasks. 

\end{abstract}


%% file: sec/1_intro.tex
\section{Introduction}
\label{sec:intro}
Dexterous manipulation with multi-fingered robotic hands has recently achieved rapid progress across diverse tasks such as grasping, flipping, and in-hand object manipulation~\cite{fu2025cordvip,li2025maniptrans,zhong2025dexgraspvla,Li2025ScalableVM,Kedia2026SimToolRealAO}. 
Despite these advances, acquiring high-quality supervision remains a central bottleneck. 
Most existing methods rely on carefully teleoperated robot demonstrations, which are costly to obtain, time-consuming, and often coupled to specific robot embodiments, camera viewpoints, and scene configurations~\cite{gervet2023act3d,haldar2023teach,wang2024equivariant,lin2025learning,zhao2023learning}. 
This bottleneck is further amplified for dexterous robotic hands, where the high-dimensional action space and complex multi-contact dynamics make demonstration collection substantially more difficult. Moreover, lack of haptic feedback and limited sampling rates can disrupt the fluidity of teleoperated demonstrations; policies trained on these demonstrations struggle to capture the fluid human motion patterns.

The aforementioned limitations motivate a shift from robot data collection toward HOI videos, which offer a more scalable and embodiment-agnostic source of supervision for dexterous skill acquisition~\cite{Li2025ScalableVM,Luo2025BeingH0VP,qin2022dexmv,Ye2025Video2PolicySU, Chen2026DexterousMP, Gupta2026LUCIDLE}. However, directly learning robotic manipulation from such videos remains highly challenging. 
Existing video-based dexterous learning methods typically model hand motion from human videos and imitate corresponding skills via reinforcement learning, but their validation largely remains confined to simulated environments or their real-world deployment lacks closed-loop perception of interaction dynamics~\cite{qin2022dexmv,mandikal2022dexvip}. We therefore propose \nbname{}, a framework that directly transfers manipulation skills from human video demonstrations to closed-loop execution on real robots. Moreover, prior works including ~\cite{Luo2025BeingH0VP,qin2022dexmv,Shaw2022VideoDexLD,Tao2025DexWildDH,li2026egolive} either treat human videos as visual priors or encode hand motion through factorized joint representations, limiting the emphasis on modeling the inter-joint coupling and temporal context that are essential for dexterous manipulation. To address this limitation, we introduce the DexHand Motion Perception Module (DMPM), which represents the hand as a directed skeletal graph, enabling the module to model kinematic dependencies and coordinated motion patterns for action generation. Leveraging large-scale HOI data, we pretrain DMPM on over 5M frames to learn transferable motion representations from a broad spectrum of behaviors.


We evaluate \nbname\,on six real-world dexterous manipulation tasks using a five-fingered robotic hand. 
Without using any real-robot demonstrations, \nbname\,achieves strong real-world performance and supports zero-shot cross-scene deployment under variations in object pose, camera viewpoint, and robot initial state. 
Extensive experiments and ablations demonstrate the effectiveness of video-derived expert demonstrations and proposed visuomotor policy. In summary, our contributions are as follows:
\begin{enumerate}
\item We present \nbname, a new framework for learning from human videos, which transfers dexterous manipulation skills from monocular human videos to real-world control without any on-robot demonstrations.
\item We introduce the DexHand Motion Perception Module (DMPM) to capture finer-grained robotic hand motion context, enhancing dexterous manipulation.
\item Through extensive real-world experiments, we demonstrate that \nbname\,substantially advances dexterous imitation learning, achieving an average success rate of up to 87.5\% and exhibiting strong generalization.
\end{enumerate}

%% file: sec/2_related_work.tex
\section{Related Work}

\subsection{Dexterous Hand Motion Representation}
Learning compact representations of high-dimensional hand motion
has emerged as an effective approach to dexterous control.
Liconti et al.~\cite{liconti2024leveraging} pretrain latent
representations of hand-motion subtrajectories for few-shot
imitation learning, while VQ-ACE~\cite{yang2024vq} compresses
human hand motions into quantized action chunks to facilitate
planning and reinforcement learning.
More recently, XL-VLA~\cite{jiang2026cross} learns a shared
latent action space across dexterous hand embodiments.  Being-H0~\cite{Luo2025BeingH0VP} investigates alternative hand-motion tokenization strategies to identify effective representations for downstream manipulation.
These studies motivate transferable motion representations for dexterous manipulation but do not explicitly account for the intrinsic coupling and strong dependencies among finger joints.
Our approach emphasizes topology-aware hand-state encoding by explicitly incorporating the robotic hand’s skeletal structure.
\subsection{Geometry-Aware Visuomotor Policies}
 Recent diffusion-based visuomotor policies integrate 3D representations into conditional action generation to improve spatial generalization. DP3~\cite{ze20243d} extracts compact geometric features from sparse point clouds, while 3D Diffuser Actor~\cite{ke20243d} integrates 3D scene features, language, and proprioception within a denoising transformer. Subsequent approaches enrich these representations with interaction and temporal structure. CordViP~\cite{fu2025cordvip} incorporates hand–object correspondences and object-centric contact maps for dexterous manipulation, whereas DP4~\cite{Liu2025SpatialTemporalAV} captures scene dynamics through a dynamic Gaussian world model. Geometric equivariance provides a complementary inductive bias. Equivariant Diffusion Policy~\cite{wang2024equivariant} enforces SO(2) symmetry in action denoising, and EquiBot~\cite{yang2024equibot} incorporates $\mathrm{SIM}(3)$ equivariance to support generalization across translations, rotations, and scales. Inspired by this approach, we employ a $\mathrm{SIM}(3)$-equivariant point-cloud encoder to extract geometric representations for policy conditioning.
\subsection{Learning from Human Videos}
Human videos have become an increasingly important source of supervision for embodied learning, particularly for imitation learning.
However, how to convert such videos into effective supervision for dexterous robotic manipulation remains an open problem. 
Representative works such as~\cite{Luo2025BeingH0VP,Li2025ScalableVM,Zheng2026EgoScaleSD,Qiu2025HumanoidP} exploit estimated human hand trajectories from large-scale video corpora to learn generalizable manipulation priors. 
While these methods benefit from broad behavioral coverage, object states are usually absent or only weakly modeled. 
Moreover, they still require robot demonstrations during post-training, and therefore do not fully eliminate the dependence on robot data.
Methods including \cite{qin2022dexmv,mandikal2022dexvip,Hsieh2025DexManLB} recover more structured hand-object interaction trajectories from videos, and use the recovered motion to reproduce corresponding dexterous skills in simulation. These methods provide a more explicit geometric bridge between human videos and robotic manipulation. Nevertheless, these methods are either evaluated only in simulation or deployed on robots by simply replaying simulation rollouts, rather than following a closed-loop mechanism that conditions action generation on real-world observations.

%% file: sec/3_method.tex
\section{Method}
\subsection{Overview}
\begin{figure*}[!h]
  \centering
  \includegraphics[width=\textwidth]{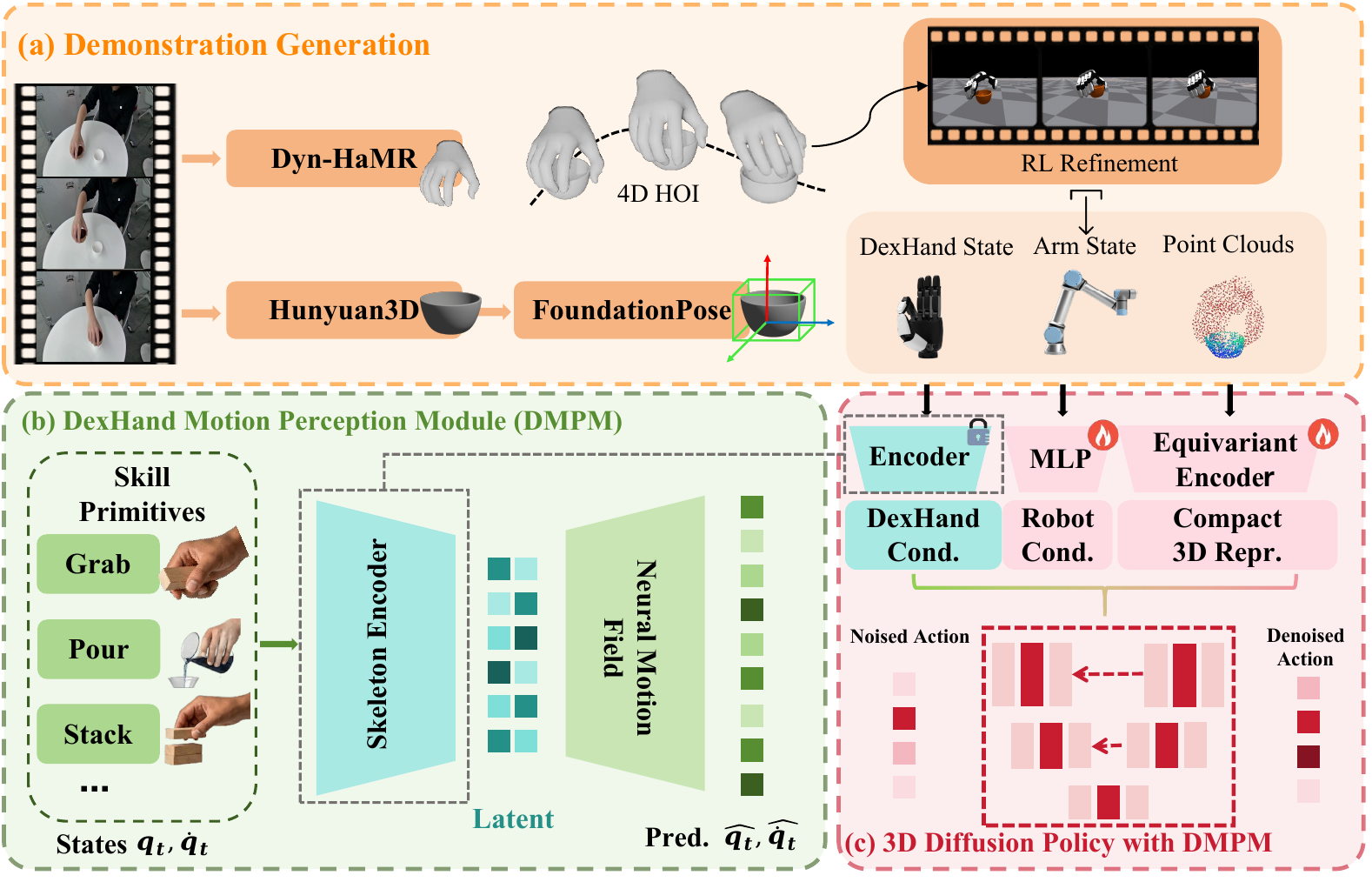}
  \caption{Overview of \nbname. (a) Demonstration acquisition. Given monocular videos, we leverage 4D HOI reconstruction and reinforcement learning to generate training data for dexterous policy learning. (b) We pretrain our VAE-based DexHand Motion Perception Module (DMPM) on large-scale motion capture data. (c) The pretrained DMPM encoder produces a compact latent representation of robotic hand motion, which conditions the diffusion process.}
  \label{fig:pipeline}
\end{figure*}

Our goal is to learn deployable real-world dexterous control directly from RGB-D HOI videos, without any real-robot demonstrations. For this purpose, we transform raw HOI videos into structured expert demonstrations for downstream policy training. Second, to provide temporal context and topology-aware motion representations for action generation, we propose a DexHand Motion Perception Module (DMPM) and incorporate its frozen encoder into our visuomotor policy. Fig.~\ref{fig:pipeline} illustrates the overall pipeline.
\subsection{Demonstration Generation}
Here, we detail how to derive expert demonstrations for imitation learning from HOI videos through human-to-robot transfer and data augmentation.

\subsubsection{Human-to-Robot Demonstration Generation}
\begin{figure*}[t]
  \centering
  \includegraphics[width=1.0\linewidth]{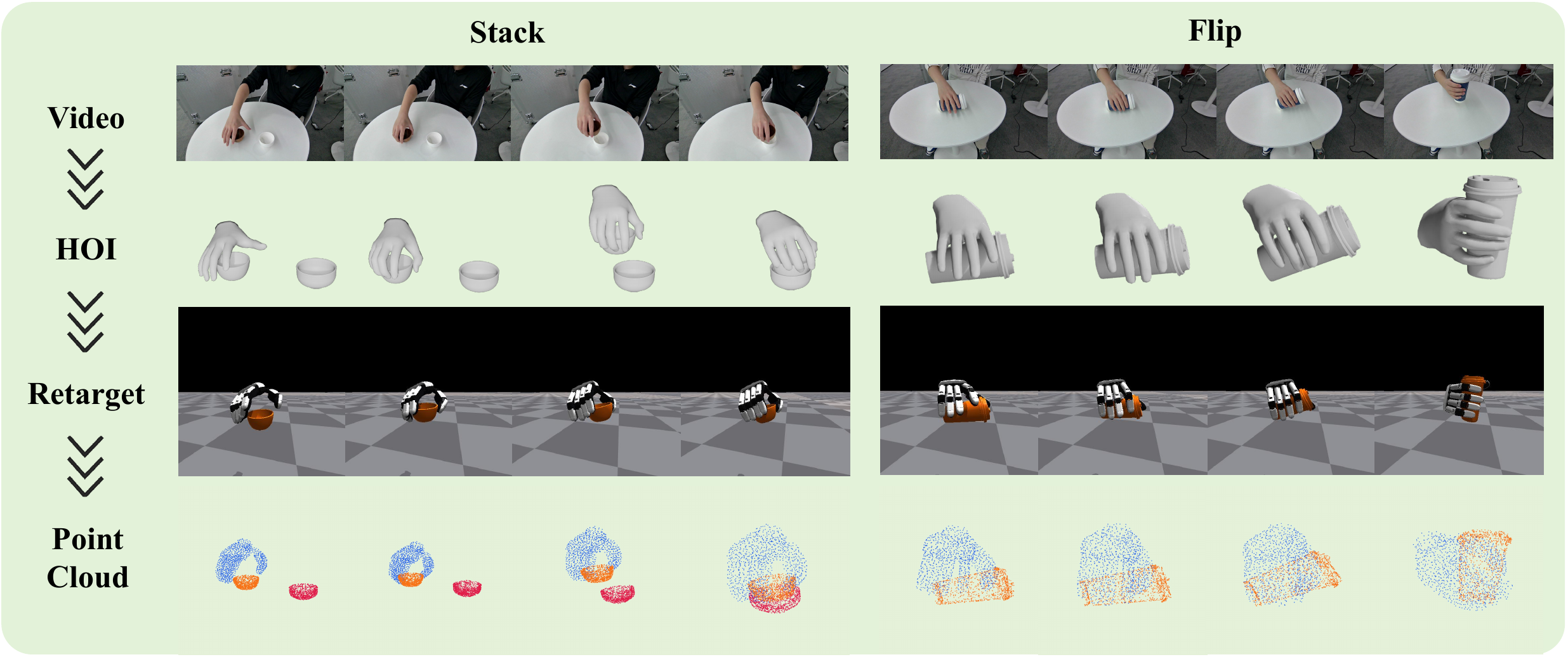}
  \caption{Visualization of data flow. We recover the full hand–object geometry from raw RGB-D videos, transfer the interaction to a robotic hand, and synthesize complete interaction point clouds for downstream policy training.}
  \label{fig:pc_generate}
  \vspace{-0.7em} 
\end{figure*}
We convert a monocular HOI video into a temporally consistent HOI representation using a video-based reconstruction pipeline following~\cite{zhang2025robowheel} and refine the trajectories of both the hand and object via a residual RL policy~\cite{li2025maniptrans} to obtain physically plausible poses. The resulting trajectories are transformed into expert demonstrations
$
\mathcal{D}^{i}=\{\mathbf{P}_{t}^{h},\mathbf{P}_{t}^{o},\mathbf{s}_{t},\mathbf{a}_{t}\}_{t=1}^{T_i}.
$ \(\mathbf{P}_{t}^{h},\mathbf{P}_{t}^{o}\in\mathbb{R}^{1024\times3}\) are the complete robotic-hand and object point clouds obtained by sampling their respective meshes. The proprioceptive state comprises the hand and arm joint
positions and velocities:
$\mathbf{s}_{t} =
[\mathbf{q}_{t}^{h};\mathbf{q}_{t}^{r};
\dot{\mathbf{q}}_{t}^{h};\dot{\mathbf{q}}_{t}^{r}]
\in\mathbb{R}^{36}$.
Correspondingly, the imitation target $\mathbf{a}_{t}$ comprises both absolute joint-position
targets and joint-velocity targets.
\subsubsection{Data Augmentation}
As original RL rollouts provide limited motion diversity, we further augment the data to achieve a broader spatial distribution while preserving the underlying interaction geometry. We randomly perturb the object pose by translating it within a $20\,\mathrm{cm}\times 20\,\mathrm{cm}$ workspace region and rotating it within a $60^\circ$ angular range. Such structured augmentation introduces controlled variations in object placement, robot initialization, and workspace configuration, thereby improving the downstream policy's generalization.
\subsection{DexHand Motion Perception Module}
\subsubsection{DMPM Architecture}
\begin{figure*}[!h]
  \centering
  \includegraphics[width=\linewidth]{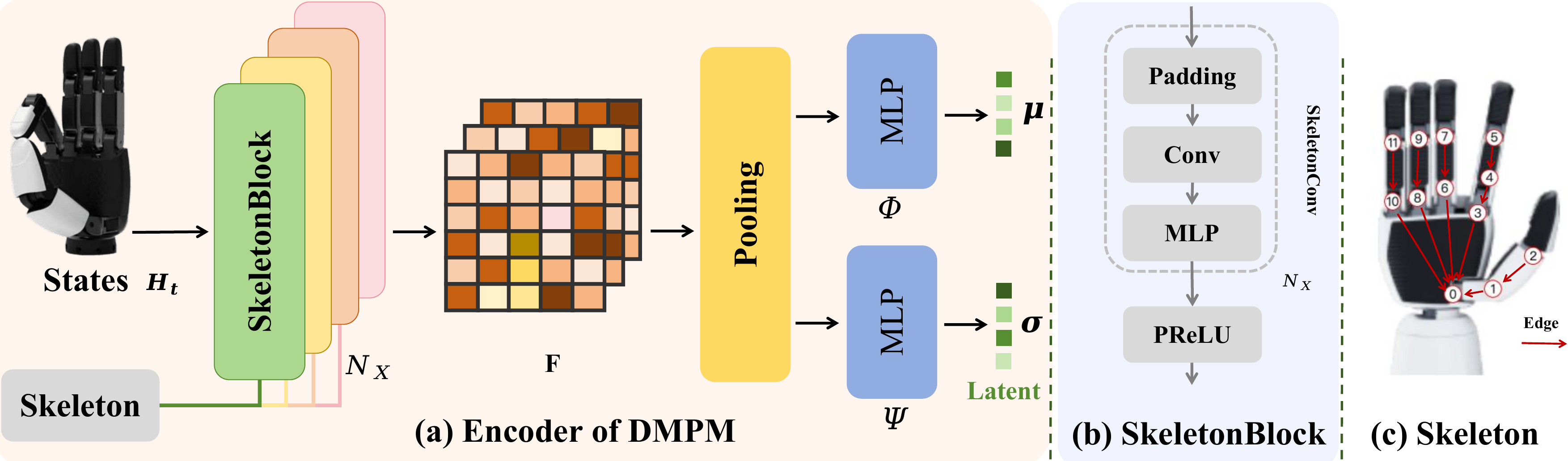}
  \caption{Details of the DMPM Encoder. (a) Our encoder leverages the articulated kinematic chains of the robotic hand to capture structured inter-joint couplings. (b) SkeletonBlock performs topology-aware residual aggregation along the kinematic graph. (c) Visualization of the skeleton. The skeleton represents the
robotic hand as a structured articulated graph, with nodes
representing joints and directed edges tracing the kinematic
hierarchy.}
  \label{fig:vaeencoder}
  \vspace{-2.0em} 
\end{figure*}
We formulate robotic hand motion perception within a latent-variable framework and instantiate it as a variational autoencoder (VAE)~\cite{Kingma2013AutoEncodingVB}, as shown in Fig.~\ref{fig:pipeline}(b). DMPM encodes robotic hand motion sequences into a compact latent representation $\mathbf{z}$ and decodes it to reconstruct the original motion sequences.

\paragraph{Encoder}
We introduce a \emph{skeleton} into the residual blocks to explicitly encode the topology of the robotic hand, as shown in Fig.~\ref{fig:vaeencoder}. Concretely, the skeleton represents the robotic hand as a structured articulated graph, with nodes representing joints and directed edges tracing the kinematic hierarchy that governs local inter-joint dependencies.
This topology restricts feature interactions to local neighborhoods on the skeleton graph, enabling the encoder to model inter-joint couplings and coordinated motion patterns beyond factorized joint representations.
\paragraph{Decoder}
We adopt the non-autoregressive Neural Motion Field (NeMF)
decoder~\cite{he2022nemf} as our VAE decoder for efficient sampling.
Our decoder $\mathcal{M}$ represents robotic hand motion as a
continuous function of time conditioned on a latent code:
$
\mathcal{M}: (t,z) \mapsto J_t,
$
where $J_t$ denotes the kinematic pose at time $t$.
Given a latent code $z$, the entire motion sequence can be sampled
in parallel by querying the decoder at different timestamps. The effectiveness of the learned latent representation is empirically evaluated through the kinematic fidelity and temporal coherence of the reconstructed motion.
\paragraph{Training Objective}
The objective function we aim to minimize is defined as:
\begin{equation}
\begin{aligned}
\min_{\theta,\phi}\quad
&\mathbb{E}_{\{\mathbf{q}_t,\dot{\mathbf{q}}_t\}_{t=1}^{T}\sim\mathcal{D}}
\Bigg[
\sum_{t=1}^{T}
\left\|
\mathbf{W}_q(\hat{\mathbf{q}}_t-\mathbf{q}_t)
\right\|_2^2
\\
&\quad+
\lambda_v
\sum_{t=1}^{T}
\left\|
\mathbf{W}_v(\hat{\dot{\mathbf{q}}}_t-\dot{\mathbf{q}}_t)
\right\|_2^2
\\
&\quad+
\lambda_{\mathrm{KL}}
\mathrm{KL}\!\left(
q_{\phi}\!\left(
\mathbf{z}\mid
\{\mathbf{q}_t,\dot{\mathbf{q}}_t\}_{t=1}^{T}
\right)
\,\|\,p(\mathbf{z})
\right)
\Bigg].
\end{aligned}
\end{equation}
Here, $\hat{\mathbf{q}}$ and $\hat{\dot{\mathbf{q}}}$ denote robotic hand poses and velocities respectively. 
The motion latent code $\mathbf{z}$ is inferred by the topology-aware encoder $q_{\phi}(\mathbf{z}\mid \{\mathbf{q}_t,\dot{\mathbf{q}}_t\}_{t=1}^{T})$ parameterized by $\phi$. The metrics $\mathbf{W}_q$ and $\mathbf{W}_v$ are empirically set to balance the importance of different joints, and $\lambda_v$ controls the trade-off between pose precision and dynamical consistency. The KL divergence regularizes the learned latent distribution toward a standard Gaussian prior $p(\mathbf{z})$.

\subsubsection{Training Data}
The data used for training DMPM are derived from  publicly available HOI datasets, including GRAB~\cite{Taheri2020GRABAD}, TCDHands~\cite{AMASS_TCDHands}, DexYCB~\cite{chao:cvpr2021}, H2O~\cite{Kwon_2021_ICCV}, ARCTIC~\cite{fan2023arctic}, HOT3D~\cite{banerjee2024hot3d}, and HoCap~\cite{wang2024hocapcapturedataset3d}. 
Inspired by~\cite{Ding2024BunnyVisionProRB}, we retarget
MANO~\cite{romero2022embodied} poses to the robotic hand by
minimizing discrepancies between their corresponding
wrist-to-fingertip vectors. The resulting pretraining corpus comprises over 5M frames spanning a broad repertoire of hand behaviors. Its scale and
diversity provide a rich foundation for learning transferable motion representations. 


\subsection{3D Diffusion Policy with DMPM}
We formulate dexterous action generation as a conditional diffusion process enhanced by the pretrained DexHand Motion Perception Module (DMPM). In our implementation, given a 32-frame robotic-hand motion sequence $H$
comprising joint positions and velocities, the frozen DMPM encoder produces a topology-aware latent embedding
$\mathbf{f}_{d}\in\mathbb{R}^{256}$. This representation captures inter-joint dependencies and temporal context in hand motion to inform future action generation.
We use separate point-cloud encoding branches $\Phi_h$ and $\Phi_o$ for the robotic-hand clouds $\mathbf{P}_t^h$ and object clouds $\mathbf{P}_t^o$, respectively. Each branch is built upon a SIM(3)-equivariant PointNet as its backbone~\cite{yang2024equibot} and outputs
rotation-equivariant and invariant latent representations,  $\Theta_R$ and $\Theta_{\mathrm{inv}}$, respectively. 
The two point-cloud embeddings are fused into the geometric condition $\mathbf{f}_g$, while an MLP maps robot proprioception to $\mathbf{f}_r$. These features are injected into the diffusion network through FiLM layers~\cite{perez2018film}.
Starting from Gaussian noise $\mathbf{a}_K$, the denoising network $\epsilon_{\theta}$ iteratively refines the noisy action for $K$ steps to recover the clean action $\mathbf{a}_0$:
\begin{equation}
\mathbf{a}_{k-1}
=
\alpha_k\Big(\mathbf{a}_k-\gamma_k\,\epsilon_{\theta}\!\left(\mathbf{a}_k,k,\mathbf{f}_g,\mathbf{f}_d,\mathbf{f}_r\right)\Big)
+\sigma_k\,\mathcal{N}(\mathbf{0},\mathbf{I}),
\end{equation}
where \(\mathcal{N}(\mathbf{0}, \mathbf{I})\) denotes Gaussian noise, and \(\alpha_k\), \(\gamma_k\), and \(\sigma_k\) are step-dependent coefficients determined by the noise scheduler. As the expert demonstrations are derived from human video with a high frame rate, our action chunk size is fixed to be 64. We use a 32-step observation history ($\approx 1$ s at 30 fps) to predict the next 32 action steps, with a total horizon of 64 steps. We train $\epsilon_{\theta}$ using the following objective,

\begin{equation}
\mathcal{L}=\mathbb{E}\!\left[
\left\|\boldsymbol{\epsilon}-\epsilon_{\theta}\!\left(
\sqrt{\bar{\alpha}_k}\mathbf{a}_0+
\sqrt{1-\bar{\alpha}_k}\boldsymbol{\epsilon},
k,\mathbf{f}_g,\mathbf{f}_d,\mathbf{f}_r\right)\right\|_2^2
\right].
\end{equation}

%% file: sec/4_exper.tex
\section{Experiments}
\subsection{Experimental Setup}
\subsubsection{Hardware Setup} Our system comprises a 6-DoF UR5 robot paired with a 12-DoF five-fingered XHAND robotic hand. A single side-mounted ORBBEC GEMINI 335 RGB-D camera provides object tracking throughout execution and we calibrate the camera
extrinsics to the robot base frame.
\subsubsection{Task Design} We design our tasks to cover commonly used human manipulation skills, including pushing, grasping, wiping, flipping, pouring, and stacking. For each task, we generate all expert demonstrations from the same source human video recorded at 30 fps; trajectory horizons range from 150 to 300 steps across tasks. We use a total of 40 demonstrations for training per task, including original RL rollouts and augmented data. We briefly describe each task as follows.
\begin{itemize}
    \item \textbf{Grasp}: Pick up the mango from the tabletop.
    \item \textbf{Pour}: Pour coffee from the bottle.
    \item \textbf{Flip}: Reorient the paper cup to an upright position.
    \item \textbf{Push}: Push the button on the tabletop.
    \item \textbf{Wipe}: Pick up the eraser and clean the whiteboard.
    \item \textbf{Stack}: Stack one cup on top of another.
\end{itemize}
\subsubsection{Baselines} We benchmark against 3D-based imitation learning algorithms (DP-3D, ACT-3D and CordViP) and a state-based policy. The state-based diffusion policy baseline is conditioned on concatenated robot proprioception and the estimated object 6-DoF pose. DP-3D and ACT-3D are the adaptations of 3D Diffusion Policy~\cite{ze20243d} and ACT~\cite{zhao2023learning}, respectively. DP-3D replaces full-scene point clouds with robotic hand–object point clouds, while ACT-3D replaces the image encoder with a point cloud encoder. We emphasize that all point-cloud observations are constructed using our framework, and the number of training point clouds is kept identical across methods. 
All models are fine-tuned for 10k iterations on each single task, with a batch size of 128, a horizon of 64, 32 observation steps, and 32 action steps.
\subsubsection{Deployment Details} Before deployment, object meshes are reconstructed using Hunyuan3D~\cite{hunyuan3d2} and scaled to metric units. FoundationPose~\cite{wen2024foundationpose} estimates the object's 6-DoF pose from RGB-D observations during inference. Points sampled from the object mesh are transformed into the
robot base frame using calibrated camera extrinsics to obtain the current object point clouds. The robotic-hand point clouds are generated from joint states through forward kinematics and mesh sampling~\cite{Wei2024DROG}. We maintain an observation buffer to provide the required input sequence, updating it after each executed action and computing required joint velocities from the buffered joint positions. Consequently, the resulting point clouds are temporally aligned with the
proprioceptive state stream. The robot controller executes the complete predicted 32-step action chunk at 8 Hz without considering velocity. After the previous actions are completed, the latest observation is used to predict the next action chunk.
\subsection{Real-world Evaluation}
\begin{figure*}[!h]
  \centering
  \includegraphics[width=0.9\linewidth]{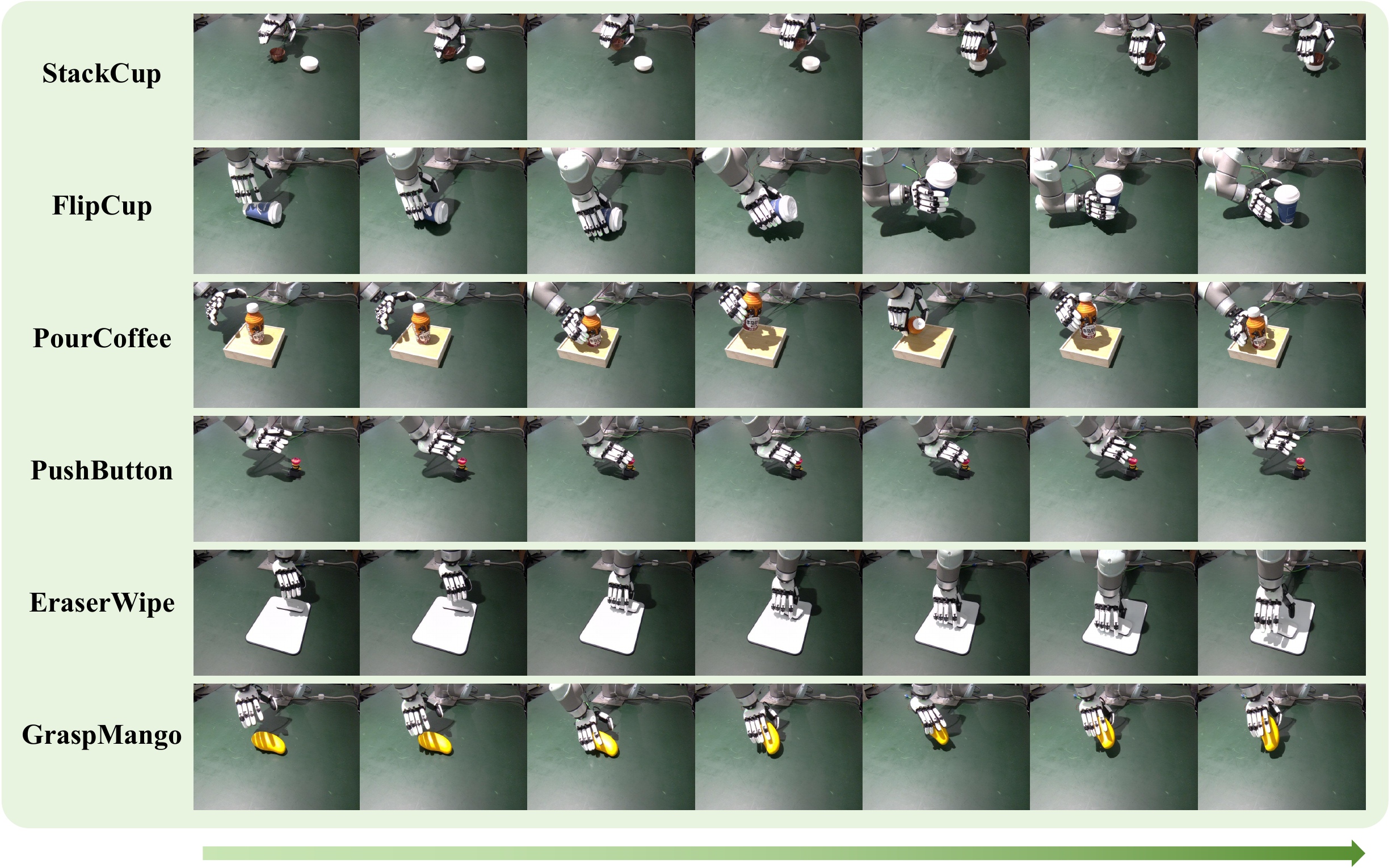}
  \caption{Real-world performance on six dexterous manipulation tasks.}
  \label{fig:real-world}
  \vspace{-1em}
\end{figure*}

\begin{table*}[!h]
    \centering
    \caption{Quantitative results on six real-world dexterous manipulation tasks.
    Each entry reports the success rate (\%) under Viewpoint-1 \textbar{} Viewpoint-2, with the final column reporting the average success rates.}
    \label{tab:task_perf_8x6_seen_unseen_with_imgs}

    \setlength{\tabcolsep}{4pt}       
    \renewcommand{\arraystretch}{1.0} 
    \fontsize{8.5pt}{10pt}\selectfont

    \begin{tabular}{@{}lccccccc@{}}
        \toprule
        \textbf{Model} &
        Grasp & Pour & Flip & Push & Wipe & Stack & Avg \\
        \midrule

        State-Based DP
        & 35.0 \textbar{} 40.0
        & 50.0 \textbar{} 40.0
        & 60.0 \textbar{} 50.0
        & 40.0 \textbar{} 35.0
        & 20.0 \textbar{} 30.0
        & 10.0 \textbar{} 25.0
        & 35.8 \textbar{} 36.7 \\

        ACT-3D
        & 40.0 \textbar{} 50.0
        & 40.0 \textbar{} 35.0
        & 65.0 \textbar{} 70.0
        & 60.0 \textbar{} 70.0
        & 25.0 \textbar{} 40.0
        & 20.0 \textbar{} 30.0
        & 41.7 \textbar{} 49.2 \\

        DP-3D
        & 35.0 \textbar{} 45.0
        & 60.0 \textbar{} 50.0
        & 60.0 \textbar{} 65.0
        & 65.0 \textbar{} 70.0
        & 60.0 \textbar{} 55.0
        & 30.0 \textbar{} 40.0
        & 51.7 \textbar{} 54.2 \\

        CordViP
        & 75.0 \textbar{} 80.0
        & 70.0 \textbar{} 75.0
        & 75.0 \textbar{} 70.0
        & 80.0 \textbar{} 80.0
        & 75.0 \textbar{} 80.0
        & 60.0 \textbar{} 65.0
        & 72.5 \textbar{} 75.0 \\

        \rowcolor{cellgray}
        \textbf{Ours}
        & \textbf{90.0} \textbar{} \textbf{80.0}
        & \textbf{85.0} \textbar{} \textbf{90.0}
        & \textbf{80.0} \textbar{} \textbf{85.0}
        & \textbf{90.0} \textbar{} \textbf{95.0}
        & \textbf{95.0} \textbar{} \textbf{90.0}
        & \textbf{85.0} \textbar{} \textbf{80.0}
        & \textbf{87.5} \textbar{} \textbf{86.7} \\

        \bottomrule
    \end{tabular}
\end{table*}
To assess the effectiveness of our framework, we perform real-world evaluations across six tasks. Experiments are conducted under two distinct camera viewpoints: Viewpoint-1 observes the workspace from the front, whereas Viewpoint-2 observes it from the side. For each viewpoint, we execute 20 trials, with the object pose and the arm’s initial state randomized at evaluation time. A trial is considered successful if the robot completes the intended task outcome within the prescribed execution horizon. The results in Tab. \ref{tab:task_perf_8x6_seen_unseen_with_imgs} show that our proposed \nbname\, outperforms all baselines across tasks, achieving average success rates of 87.5\% and 86.7\% under two viewpoints. 
These remarkable results reflect two complementary successes. First, leveraging precise HOI reconstruction, our framework achieves direct transfer from monocular videos to real-world control, without any real-robot demonstrations. Second, our 3D diffusion policy enhanced with DMPM produces more executable and feasible actions. 
Fig. \ref{fig:real-world} visualizes real-world deployment.

\begingroup
\setlength{\columnsep}{6pt}   
\setlength{\intextsep}{4pt}   

Beyond its strong in-distribution (ID) performance, \nbname\, also exhibits robust generalization to unseen objects and visually diverse environments (please refer to the accompanying video for visualizations). 
\begin{wraptable}{r}{0.55\columnwidth}
    \centering
    \scriptsize
    \setlength{\tabcolsep}{2pt}
    \renewcommand{\arraystretch}{0.95}

    \resizebox{\linewidth}{!}{%
        \begin{tabular}{lccccc}
            \toprule
            \multirow{2}{*}{\textbf{Model}} &
            \multicolumn{2}{c}{\textbf{Pour}} &
            \multicolumn{2}{c}{\textbf{Stack}} &
            \multirow{2}{*}{\textbf{Avg.}} \\

            \cmidrule(lr){2-3}
            \cmidrule(lr){4-5}

            & \textbf{Seen}
            & \textbf{Unseen}
            & \textbf{Seen}
            & \textbf{Unseen}
            & \\

            \midrule
            DP-3D
            & 60.0 & 55.0
            & 30.0 & 30.0
            & 43.8 \\

            CordViP
            & 70.0 & 65.0
            & 60.0 & 55.0
            & 62.5 \\

            \textbf{Ours}
            & \textbf{85.0} & \textbf{85.0}
            & \textbf{85.0} & \textbf{85.0}
            & \textbf{85.0} \\

            \bottomrule
        \end{tabular}%
    }

    \caption{Success rates (\%) on seen and unseen objects.}
    \label{tab:gen_unseen_obj_2task}
    \vspace{-1em}
\end{wraptable}
As reported in Tab.~\ref{tab:gen_unseen_obj_2task}, we benchmark DP-3D, CordViP, and our method on the Pour and Stack tasks using both seen and unseen objects. Compared with the baselines, \nbname\, demonstrates markedly stronger adaptability to variations in object shape, maintaining higher success rates under object-level distribution shifts.
Furthermore, Tab.~\ref{tab:gen_environment} notably highlights another key advantage of our framework. By focusing on interaction-centric geometry, we effectively decouple the policy from background appearance and our policy delivers stronger spatial perception and better performance.
\par
\endgroup

\begin{table*}[!h]
  \centering
  \scriptsize
  \setlength{\tabcolsep}{5pt}
  \renewcommand{\arraystretch}{1.05}

  \resizebox{\linewidth}{!}{%
  \begin{tabular}{@{}l*{17}{c}@{}}
    \toprule
    \multirow{2}{*}{\textbf{Model}} &
    \multicolumn{4}{c}{\textbf{Pour}} &
    \multicolumn{4}{c}{\textbf{Stack}} &
    \multicolumn{4}{c}{\textbf{Grasp}} &
    \multicolumn{4}{c}{\textbf{Push}} &
    \multirow{2}{*}{\textbf{Avg.}} \\
    \cmidrule(lr){2-5}
    \cmidrule(lr){6-9}
    \cmidrule(lr){10-13}
    \cmidrule(lr){14-17}
    & \textbf{Env-1} & \textbf{Env-2} & \textbf{Env-3} & \textbf{Cluttered}
    & \textbf{Env-1} & \textbf{Env-2} & \textbf{Env-3} & \textbf{Cluttered}
    & \textbf{Env-1} & \textbf{Env-2} & \textbf{Env-3} & \textbf{Cluttered}
    & \textbf{Env-1} & \textbf{Env-2} & \textbf{Env-3} & \textbf{Cluttered}
    & \\
    \midrule
    DP-3D
    & 60.0 & 50.0 & 45.0 & 30.0
    & 30.0 & 25.0 & 40.0 & 20.0
    & 35.0 & 25.0 & 25.0 & 20.0
    & 65.0 & 60.0 & 55.0 & 45.0
    & 39.4 \\

    CordViP
    & 70.0 & 65.0 & 65.0 & 55.0
    & 60.0 & 55.0 & 50.0 & 40.0
    & 75.0 & 70.0 & 75.0 & 60.0
    & 80.0 & 70.0 & 75.0 & 65.0
    & 64.4 \\

    \textbf{Ours}
    & \textbf{85.0} & \textbf{85.0} & \textbf{80.0} & \textbf{70.0}
    & \textbf{85.0} & \textbf{85.0} & \textbf{90.0} & \textbf{70.0}
    & \textbf{90.0} & \textbf{85.0} & \textbf{85.0} & \textbf{75.0}
    & \textbf{90.0} & \textbf{90.0} & \textbf{85.0} & \textbf{80.0}
    & \textbf{83.1} \\
    \bottomrule
  \end{tabular}%
  }
  \caption{Generalization of \nbname\,to different environments (Env.) (Success Rate \%).}
  \label{tab:gen_environment}
  \vspace{-0.5em}
\end{table*}

\subsection{Ablation Study}
\subsubsection{Ablation of DexCast}
We perform fine-grained ablation studies across six tasks to disentangle the individual contributions of the SIM(3)-equivariant architecture and the DMPM. The results displayed in Tab. ~\ref{tab:success_tasks_ablation} show that these two components are effective on their own and complementary when combined, directly supporting higher success rates. We train \nbname\,from scratch to confirm that DMPM's gains largely stem from large-scale pretraining. Furthermore, our augmentation strategy is critical to achieving high task success rates under our randomized evaluation protocol, as removing it leads to a 61.7\% reduction in success rate.

\begin{table}[!h]
    \centering

    {%
    \scriptsize
    \setlength{\tabcolsep}{1.5pt}
    \renewcommand{\arraystretch}{1.08}

    \begin{tabular*}{\linewidth}{@{\extracolsep{\fill}}lccccccc@{}}
        \toprule
        \textbf{Variant} &
        \textbf{Grasp} &
        \textbf{Pour} &
        \textbf{Flip} &
        \textbf{Push} &
        \textbf{Wipe} &
        \textbf{Stack} &
        \textbf{Avg.} \\
        \midrule

        DexCast w/o augmented data
        & 30.0 & 40.0 & 20.0 & 25.0 & 25.0 & 15.0
        & 25.8 \\

        DexCast w/o SIM(3) \& DMPM
        & 60.0 & 70.0 & 60.0 & 75.0 & 75.0 & 60.0
        & 66.7 \\

        DexCast w/o DMPM
        & 80.0 & 80.0 & 70.0 & 85.0 & 85.0 & 70.0
        & 78.3 \\

        DexCast w/o SIM(3)
        & 80.0 & 85.0 & 75.0 & 80.0 & 85.0 & 80.0
        & 80.8 \\

        DexCast (from scratch)
        & 80.0 & 85.0 & 60.0 & 90.0 & 85.0 & 80.0
        & 80.0 \\

        \textbf{DexCast (Full)}
        & \textbf{90.0}
        & \textbf{85.0}
        & \textbf{80.0}
        & \textbf{90.0}
        & \textbf{95.0}
        & \textbf{85.0}
        & \textbf{87.5} \\

        \bottomrule
    \end{tabular*}
    }

    \caption{Ablation studies of DexCast. (Success Rate \%).}
    \label{tab:success_tasks_ablation}
    \vspace{-0.7em}
\end{table}

\subsubsection{Ablation of the DexHand Motion Perception Module}
\label{DMPM_ablations}
We conduct comprehensive ablation studies to systematically analyze the DMPM and inform the final design of DMPM.


\begin{figure}[!t]
  \centering
\includegraphics[width=\columnwidth]{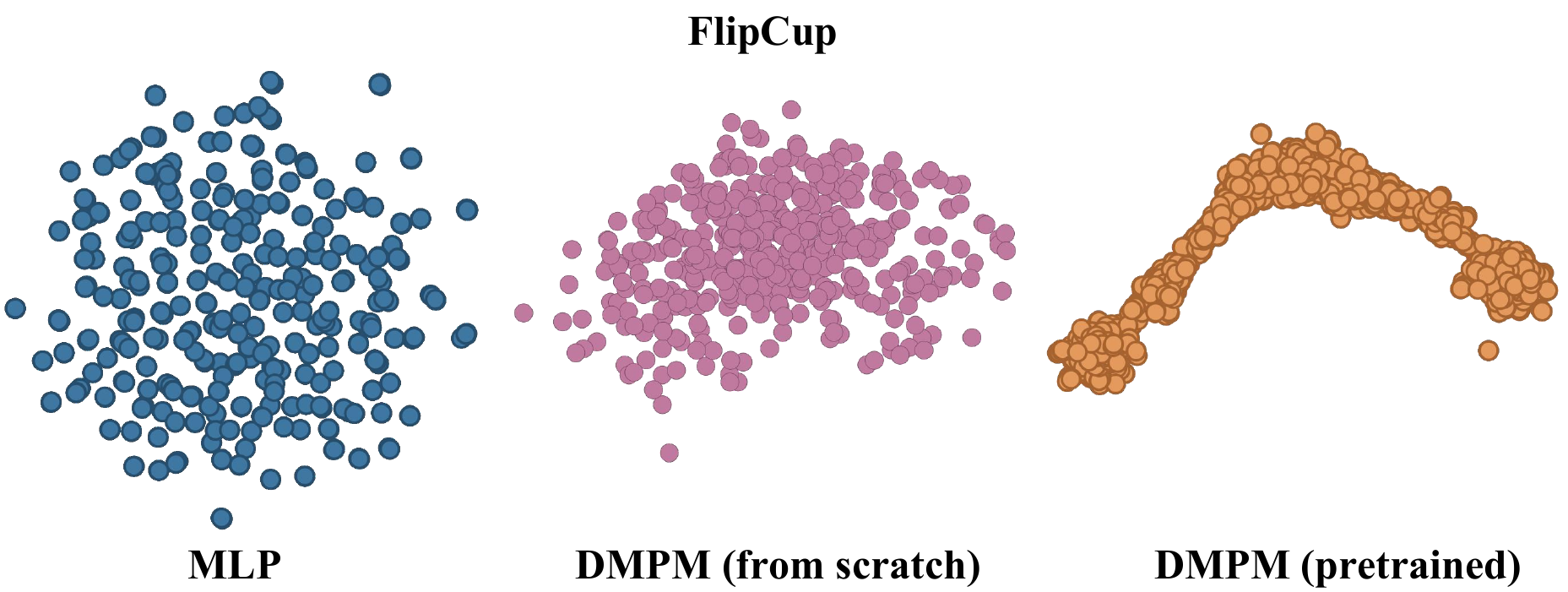}
  \caption{UMAP visualization of hand-state representations
  encoded by DMPM and an MLP baseline.}
  \label{fig:umap}
  \vspace{-2em}
\end{figure}
To assess the granularity of the learned motion representations, we compare DMPM with the MLP-based hand-state encoders commonly adopted in prior work. As shown in Fig.~\ref{fig:umap}, DMPM captures more nuanced patterns in task-specific hand motion, which become more pronounced after pretraining. This comparison suggests that our proprioceptive encoder more effectively models high-dimensional robotic hand motion, providing more informative state representations to condition the policy for action generation.

\begin{figure}[!h]
  \centering
  \includegraphics[width=1.0\linewidth]{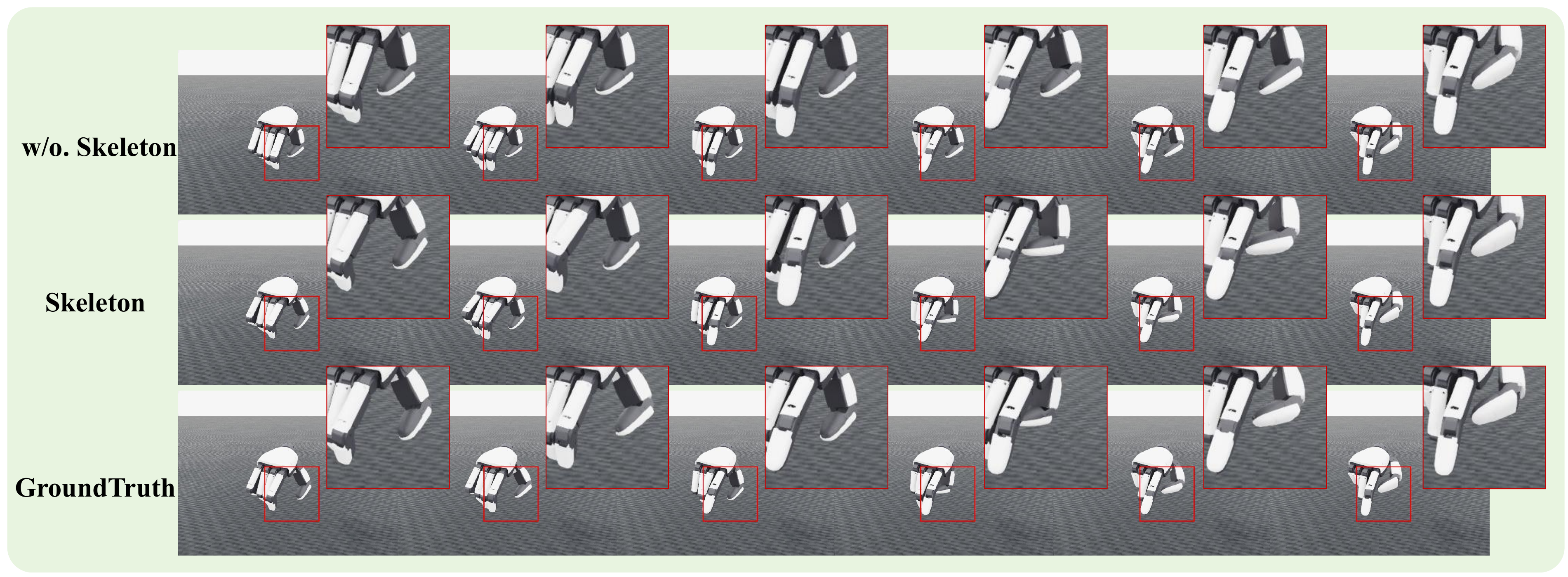}
  \caption{Visualization of DexHand motion recovery for two DMPM architectures. The DMPM with Skeleton produces higher-fidelity reconstructions.}
  \label{fig:motion_recovery_nonoise}
  \vspace{-1em}
\end{figure}

We further compare encoder variants with and without explicit skeletal modeling. To assess the performance of different variants, we evaluate motion reconstruction using Mean Per-Joint Position Error (MPJPE), Tip Mean Per-Joint Position Error (TMPJPE), Pose Consistency, and Hand Jitter. MPJPE and TMPJPE quantify Cartesian position errors across all hand joints and the fingertip subset, respectively. Pose Consistency assesses the angular reconstruction fidelity, while Hand Jitter measures temporal smoothness using joint angular acceleration. As shown in Table~\ref{tab:dmp_recon_compare}, incorporating the kinematic
structure improves reconstruction accuracy, albeit with a slight reduction in temporal smoothness. Fig.~\ref{fig:motion_recovery_nonoise} provides a visual comparison of the two variants.
\begin{table}[t]
    \centering

    {%
    \scriptsize
    \setlength{\tabcolsep}{1pt}
    \renewcommand{\arraystretch}{1.08}

    \begin{tabular*}{\linewidth}{@{\extracolsep{\fill}}l*{8}{c}@{}}
        \toprule
        \multirow{2}{*}{\textbf{Metric}}
        & \multicolumn{4}{c}{\textbf{w/o Skeleton}}
        & \multicolumn{4}{c}{\textbf{Skeleton}} \\

        \cmidrule(lr){2-5}
        \cmidrule(lr){6-9}

        & \textbf{16} & \textbf{32}
        & \textbf{64} & \textbf{128}
        & \textbf{16} & \textbf{32}
        & \textbf{64} & \textbf{128} \\
        \midrule

        TMPJPE (mm) $\downarrow$
        & 11.497 & 9.328 & 8.445 & 13.141
        & 4.005 & 3.691 & 4.540 & 12.345 \\

        MPJPE (mm) $\downarrow$
        & 3.355 & 2.722 & 2.364 & 3.835
        & 1.150 & 1.073 & 1.316 & 3.653 \\

        Pose Consistency (deg) $\downarrow$
        & 2.125 & 1.811 & 1.505 & 2.096
        & 0.776 & 0.699 & 0.863 & 1.941 \\

        Hand Jitter (deg/s$^{2}$) $\downarrow$
        & 0.133 & 0.215 & 0.215 & 0.248
        & 0.320 & 0.290 & 0.255 & 0.288 \\

        \bottomrule
    \end{tabular*}
    }

    \caption{Ablation studies on skeleton architecture and sequence length. Clip length \(T\) varies from 16 to 128 frames.}
    \label{tab:dmp_recon_compare}
    \vspace{-2.5em}
\end{table}

As for the ablations of clip length and latent-code dimensionality, Tab.~\ref{tab:dmp_recon_compare} shows that the reconstruction quality of the skeleton-based DMPM peaks at a moderate clip length ($T=32$) and degrades as $T$ increases or decreases. Using the skeleton-based architecture, we further evaluate latent dimensionalities ranging from $32$ to $2048$ (Tab.~\ref{tab:dmp_latent_compare}). The results reveal that increasing the latent size improves reconstruction up to a peak at $256$. Larger latent codes provide diminishing returns and can even degrade accuracy while generally producing progressively smoother trajectories.

\begin{table}[t]
    \centering

    {%
    \scriptsize
    \setlength{\tabcolsep}{1.5pt}
    \renewcommand{\arraystretch}{1.08}
    \setlength{\extrarowheight}{0.2pt}

    \begin{tabular*}{\linewidth}{@{\extracolsep{\fill}}l*{7}{c}@{}}
        \toprule
        \textbf{Metric}
        & \textbf{32}
        & \textbf{64}
        & \textbf{128}
        & \textbf{256}
        & \textbf{512}
        & \textbf{1024}
        & \textbf{2048} \\
        \midrule

        TMPJPE (mm) $\downarrow$
        & 6.623 & 3.370 & 3.075 & 3.024
        & 3.402 & 3.691 & 4.211 \\

        MPJPE (mm) $\downarrow$
        & 1.877 & 0.967 & 0.893 & 0.881
        & 0.983 & 1.073 & 1.221 \\

        Pose Consistency (deg) $\downarrow$
        & 1.237 & 0.639 & 0.622 & 0.581
        & 0.656 & 0.699 & 0.794 \\

        Hand Jitter (deg/s$^{2}$) $\downarrow$
        & 0.366 & 0.395 & 0.388 & 0.346
        & 0.312 & 0.290 & 0.298 \\

        \bottomrule
    \end{tabular*}
    }

    \caption{Latent Dimension Ablation. Performance of DMPM under different latent dimensions.}
    \label{tab:dmp_latent_compare}
    \vspace{-2.5em}
\end{table}

%% file: sec/5_conclusion.tex
\section{Conclusion}

 Our framework, \nbname, establishes a new pathway for learning dexterous skills directly from human video demonstrations. Leveraging a 4D HOI representation and RL refinement, \nbname\, produces training data for policy learning, replacing traditional, labor-intensive demonstration collection. We introduce a DexHand Motion Perception Module (DMPM) to provide a compact, motion-aware conditioning representation for dexterous manipulation. Our 3D diffusion policy enhanced with DMPM demonstrates strong real-world dexterous manipulation performance.
 
 \textbf{Limitations.} Our current 4D HOI reconstruction is limited to rigid objects and remains susceptible to tracking failures under rapid motion and severe occlusion, potentially compromising policy learning and real-world execution. DMPM also remains embodiment-specific, requiring separate skeleton graph design and training for different robotic hands.
\vspace{0.5em}